# Unsupervised image segmentation by Global and local Criteria Optimization Based on Bayesian Networks

**Mohamed Ali Mahjoub**[1] and **Mohamed Mhiri**[2]

[1,2]Research Unit SAGE, National Engineering School of Sousse
Technology Park, 4054, Sahloul, Sousse
University of Sousse - Tunisia
Email : medali.mahjoub@ipeim.rnu.tn, mhirimohamed@hotmail.com

**ABSTRACT**

*Today Bayesian networks are more used in many areas of decision support and image processing. In this way, our proposed approach uses Bayesian Network to modelize the segmented image quality. This quality is calculated on a set of attributes that represent local evaluation measures. The idea is to have these local levels chosen in a way to be intersected into them to keep the overall appearance of segmentation. The approach operates in two phases: the first phase is to make an over-segmentation which gives superpixels card. In the second phase, we model the superpixels by a Bayesian Network. To find the segmented image with the best overall quality we used two approximate inference methods, the first using ICM algorithm which is widely used in Markov Models and a second is a recursive method called algorithm of model decomposition based on max-product algorithm which is very popular in the recent works of image segmentation. For our model, we have shown that the composition of these two algorithms leads to good segmentation performance.*

## 1. INTRODUCTION

Despite the large number of works devoted to the field of image segmentation, algorithms are limited in the sense of a general approach to the problem. Indeed, most of the segmentation algorithms are designed to solve a specific problem. And most often, they do not separate the specific properties of the problem. In this context, we propose a new segmentation method applicable to various problems using Bayesian networks that are recommended in such cases where it is faced with problems of uncertainty of results.

This approach tries to find the segmented image with the best overall quality which is based on a set of attributes that represent local measures to assess local levels of this segmentation. These evaluation measures are actually degrees of verification of a set of predicates and terms which are previously defined. The notion of locality in turn, is expressed using a set of superpixels which are the results of over-segmentation. The philosophy of this approach is to choose these local levels in a way to be connected together to keep the overall appearance of segmentation.

Segmentation approaches can be divided into two groups: The probabilistic approach which makes the segmentation problem as a stochastic optimization problem and the deterministic approaches which include clustering contour methods and region methods. These probabilistic approaches are also divided into two groups: one group uses graphical models such as Markov random fields and Bayesian networks to model the joint probability law of linked entities image (Brunel, Nicolas 2008, K.E. Avrachenkov 2002) . Another group uses directly the laws of probability entities image without using graphical models. this includes discriminative approaches (S. Zheng 2007) leading to measures like contrast, texture and generative approaches (S. Khan 2001) which lead to a modeling pieces of the picture and hybrid approaches that combine template and discriminative generative model (Z. Tu 2003). Experience shows that the "best" segmentation techniques are those that diverse types of information and which solve this problem in a probabilistic way, that is the case with probabilistic graphical models, we provide an efficient way to model the different types of entities image, their inaccuracies and the various constraints that connect them and their uncertainties.

The plan of this paper is as follows: We begin by recalling in Section 2 an overview of Bayesian networks and the main work in image segmentation. Section 3 provides the basic principles of segmentation which our model is based on and the proposed model and the inference algorithms used in the pixels classification are illustrated in section 4. Finally, experiments are detailed in Section 5. This paper concludes in Section 6.

## 2. BAYESIAN NETWORK FOR IMAGE SEGMENTATION

Applied in image analysis, Bayesian networks have several advantages. They effectively represent the causal relationships between the various entities of image (F.V Jensen 2001); several studies demonstrate the effectiveness and performance of this model in the field of image segmentation. We present in this section some basic concepts of this model then we will look at some work using this model and reflect a very high efficiency.

A Bayesian network is a causal graph which has been associated a probabilistic representation. This representation allows making quantitative reasoning about causality that can be done inside the graph (P.Naim 2005). Thus, a Bayesian network consists of a directed acyclic graph whose nodes are random variables which may have a discrete number of possible states or whose values are continuous and according to a continuous distribution, and a set of local probability distributions which are the network parameters. For each node we have a table of conditional probabilities when the table depends on time it is called dynamic Bayesian network. The use of Bayesian networks is essential to calculate the conditional probabilities of events connected to each other by relations of cause and effect. This use is called inference.

A Bayesian network is defined by:

- ✓ A directed graph G= (V, E) without circuit, where V is the set of nodes of G, and E the set of edges of G;
- ✓ A finite probability space $(\Omega, Z, p)$ ;
- ✓ A set of random variables associated with the nodes of the graph defined on $(\Omega, Z, p)$, such as :

$$p(V_1, V_2, \ldots, V_n) = \prod_{i=1}^{n} p(V_i | C(V_i))$$

Where $C(V_i)$ is the set of causes (parents) of $V_i$ in G.

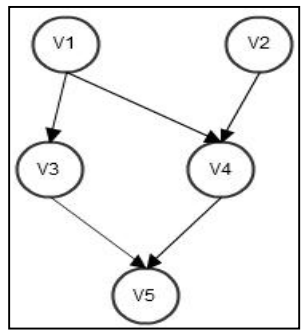

Fig .1: Bayesian network example

For image segmentation, the BN model can be used to represent a causally prior knowledge of some statistical relationships between different image entities to conclude the best possible segmentation using approximate or exact inference algorithm, which seeks the maximum of an associate objective function. In this context we will present in this section recently published approaches that use this type of model.

Lei Zhang, and Qiang Ji (Lei Zhang 2010), have proposed a Bayesian Network Model for Both Automatic (Unsupervised) and Interactive (Supervised) image segmentation. They Constructed a Multilayer BN from the over-segmentation of an image, which find object boundaries according to the measurements of regions, edges and vertices formed in the over-segmentation of the image and model the relationships among the superpixel regions, edge segment, vertices, angles and their measurements. For Automatic Image Segmentation after the construction of BN model and belief propagation segmented image is produced. For Interactive Image Segmentation if segmentation results are not satisfactory then by the human intervention active input selection are again carried out for segmentation.

Eric N. Mortensen and Jin Jia (E.N. Mortensen 2006) proposed a two layer BN model for image segmentation, which captures the relationships between edge segments and their vertices. Given a user input seed path, they use minimum path spanning tree graph search to find the most likely object boundaries. They also encode a statistical similarity measure between the adjacent regions of an edge into its a priori probability therefore implicitly integrating region information.

Kittipat Kampa, Duangmanee Putthividhya and Jose C. Principe (Kittipat, Kampa 2011) design a probabilistic unsupervised framework called Irregular Tree Structure Bayesian Network (ITSBN). The ITSBN is made according to the similarity of image regions in an input image. ITSBN is a Directed acyclic graph (DAG) with two disjoint sets of random variables hidden and observed. The original image is over-segmented in multiscale hierarchical manner then they extracted features from the input image corresponding to each superpixel. According to these superpixels ITSBN is built for each level. After applying the learning and inference algorithms the segmented image is produced.

Costas Panagiotakis, Ilias Grinias, and Georgios Tziritas (Costas Panagiotakis 2011) proposed a framework for image segmentation which uses feature extraction and clustering in the feature space followed by flooding and region merging techniques in the spatial domain, based on the computed features of classes. A new block-based unsupervised clustering method is introduced which ensures spatial coherence using an efficient hierarchical tree equipartition algorithm. They divide the image into different-different blocks based on the feature description computation. The image is partitioned using minimum spanning tree relationship and mallows distance. Then they apply K-centroid clustering algorithm and Bhattacharya distance and compute the posteriori distributions and distances and perform initial labelling. Priority multiclass flooding algorithm is applied and in the end regions are merged so that segmented image is produced.

### 3. PROPOSED MODEL PRINCIPALS

In the framework of statistical modeling, where Y represents the interpretation of the observed data of the image denoted by X, obtaining Y can be posed as an optimization problem of determining the best possible interpretation, formally the desired result is:

$$Y^* = argmax\ P\ (Y_k|X)$$

$$= argmax\ P\ (X|Y_k)\ *\ P\ (Y_k)\ /P(X)$$

$$= argmax\ P\ (X|Y_k)*P\ (Y_k)$$

The probability P ($Y_k$|X) can be seen as a function that represents the overall quality of the interpretation $Y_k$, It is composed of two terms: a term for data that expresses the likelihood between the X and $Y_k$ and another term expresses the quality of the interpretation.

In our approach we sought to model P ($Y_k$|X) in a simple and a causal way using Bayesian network. This quality is modeled by two terms, a term expresses the probability to have, and another term, expresses the evaluation quality of this interpretation $Y_k$. We use the superpixels as basic elements to avoid any kind of noise and to reduce the processing time, knowing that superpixel is a polygon part of an image containing several pixels are almost the same characteristics, in this case the evaluation quality is calculated according to a set of local features which are measures of small regions evaluation constituting the segmented image, the idea is to choose these regions to be intersected with each other to keep the overall appearance of segmentation. We seek to find an image generally well segmented not locally. These measures are verification degrees of a set of predicates.

The second term expresses the probability to have the interpretation $Y_k$ it can be seen as the local likelihoods between the superpixels of the segmented image and those of the image. The structure of the proposed model can be considered in the following diagram (Fig.4):

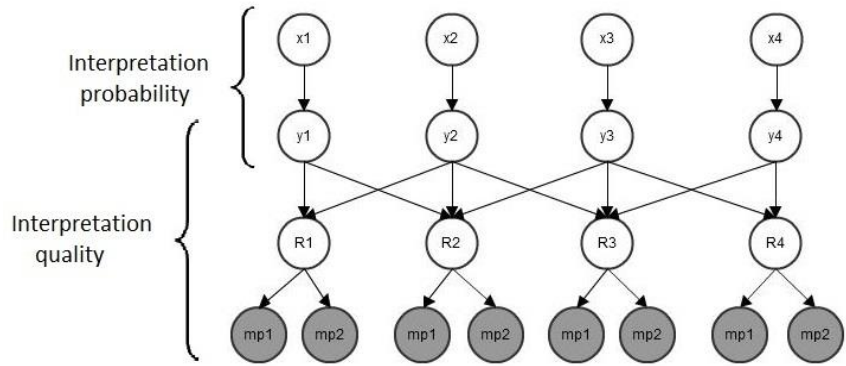

Fig.4: Proposed network model

Giving an image to be segmented, it is first necessary to over-segmentation to find superpixels, In our model, superpixels are represented by a family of random variables {$X_i$} where each variable takes its value in {0 ... 255} corresponds to the average intensity of its pixels, and for each superpixel $X_i$ we will define a random variable $Y_i$ representing this superpixel in the segmented image and takes the value of {1 .. k}, where k represents the number of cluster centers results of a clustering algorithm as k-means.

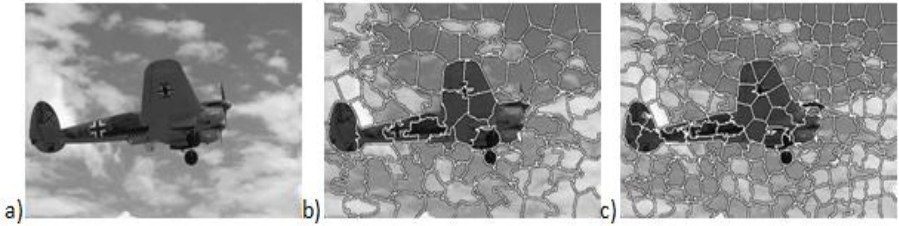

Fig.5: Over-segmentation results a) Original image b)100 superpixels c) 200 superpixels

Thus we define for each random variable $Y_i$ a random variable $R_i$ represents the region contains this superpixel and its neighboring superpixels, this region can be seen as $R_i$ =SR1 ∪ $Y_i$ ∪ SR2 (Fig.7), where SR1 represents the $R_i$ sub-region which has the same class as $Y_i$ and SR2 is the complementary sub-region to SR1 in $R_i$. Each $R_i$ variable has a value in {0...255} represents the mean of (SR1 ∪ $Y_i$) pixels intensities.

For each variable $R_i$ we will define a list of continuous random variables $\{mp_1, mp_2, ..., mp_n\}$ represent the $R_i$ verification degrees of predicates $\{p_1, p_2, ..., p_n\}$ which are previously defined. A predicate p seeks to express the homogeneity of $Y_i$ with SR1 and the contrast of $Y_i$ with SR2.

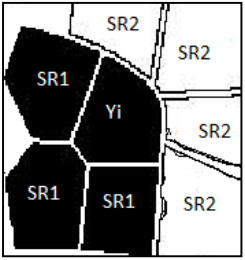

Fig.7: Example $R_i$ region

In this case, the segmentation problem can be done by finding the most probable combination Y* (MAP) of all Y using the proposed model then:

$$Y^* = \mathrm{argmax}_Y \ P\ (Y|X, R, mp)$$
$$= \mathrm{argmax}_Y \ P\ (Y, X, R, mp)$$

$$= \text{argmax}_Y \prod_{i=1}^{n} P(Y_i | X_i) * \prod_{i=1}^{n} P(R_i | pa(R_i)) * \prod_{i=1}^{n} \prod_{j=1}^{m} P(mp_j | R_i) \quad (3)$$

In equation (3), the product $\prod_{i=1}^{n} P(Y_i | X_i)$ expresses the probability to have the interpretation Y defined by $\{Y_i\}$ and knowing the image X defined by $\{X_i\}$ where $P(Y_i | X_i)$ expresses the likelihood between the superpixel $Y_i$ in the segmented image and the $X_i$ in the image, in this case $\sum_j P(Y_i = C_j | X_i = V) = 1$.

On the other hand the term $\prod_{i=1}^{n} P(R_i | pa(R_i)) * \prod_{i=1}^{n} \prod_{j=1}^{m} P(mp_j | R_i)$ expresses the evaluation quality of the interpretation Y where $\prod_{j=1}^{m} P(mp_j | R_i)$ expresses the verification degrees of region $R_i$ by m predicates and $P(R_i | pa(R_i))$ represents the probability that the sub-region (SR1 ∪ $Y_i$) be as the class of $Y_i$ where:

$$P(R_i | pa(R_i)) = P(R_i | Y_i, \text{vosinage}(Y_i))$$
$$= P(R_i = V | Y_i = C); \{V \text{ is the average pixels intensity of SR1} \cup Y_i\}.$$

This probability $P(R_i | pa(R_i))$ is calculated by a set of normal distributions, where for each class we associate the normal distribution with expectation μ and standard deviation σ, μ is the center of this class and σ a parameter initialize by ourselves represents its variance, (Fig.8) shows an example.

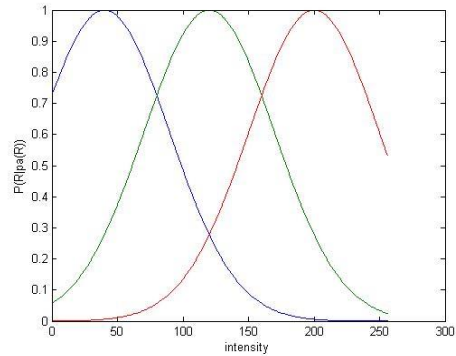

Fig.8: Examples of three normal distributions of expectations [40, 120, 200] and deviation σ=50 to calculate $P(R_i | pa(R_i))$

The probability $P(Y_i | X_i)$ is calculated from the some normal distributions, but after a normalization to check the condition $\sum_j P(Y_i = C_j | X_i = V) = 1$. An example is shown in the following (fig.9):

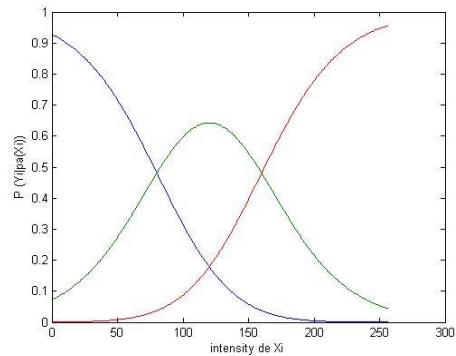

Fig.9: Examples of three normal distributions of expectations [40, 120, 200] and deviation σ=50 to calculate $P(Y_i | X_i)$

The probability $P(mp|R_i)$ of a region $R_i$, if the region $R_i$ checks the predicate then $P(mp|R_i)=0.8$ else $P(mp|R_i)=0.2$.

## 4. INFERENCE ALGORITHMS

The search for the most probable Y in the proposed model can be made either by an exact method gives an optimal solution which corresponds to the maximum of P (Y, X, R, mp) or by an approximate method gives an approximate solution corresponds to a local maximum of P (Y, X, R, mp). In our case we cannot use exact method *seen its time treatment complexity,* For this, we use two approximate inference methods express good performance, the first uses the relaxation algorithm ICM (Iterated Conditional Mode) which is widely used in the Markov model and the second is a recursive method based on 'max-product algorithm' very popular in the recent segmentation work by Bayesian networks called algorithm of model decomposition.

a. *ICM algorithm (Besag, J. 1986)*

ICM (Iterated Conditional Mode) is a deterministic iterative algorithm which converges quickly to the solution, very used in Markov random fields. In our case we adapt this algorithm in Bayesian network inference. Starting from an initial configuration of $\{Y_i\}$, the principle of this algorithm is to scan all superpixels $\{Y_i\}$, in a predetermined order, and update its classes, purpose to increase the local likelihood for each superpixel then overall interpretation probability with each iteration. Changing a class superpixel affects the local probability calculation of its neighbors, so it is necessary to repeat this scan until stable results or when stopping criterion is met. This stopping criterion can be, for example, the number of superpixel modified or the number of iterations performed generally, the threshold is around 10% of total number of superpixels. With this method, the final segmentation quality depends heavily from the initial configuration, this algorithm is therefore as follows:

**Input :**
- Image to segment, $[C_1,.., C_k]$: list of class centers.

**Output :**
- Segmented image.

**Algorithm :**

i. Initialization, it is to provide a classified image to the ICM algorithm.
ii. From this initial configuration, the following steps are performed for each superpixel ( $Y_i$ ) :
   1. For each class $C_j$ we calculate :
   $$P = P(Y_i = C_j \mid X_i) * P(R_i \mid pa(R_i)) * \prod_{j=1}^{m} P(mp_j \mid R_i).$$
   2. Find the class with the maximum probability.
iii. We retain the class C$_{max}$, find in step 2), for the superpixel $Y_i$
iv. If the changes number is greater than a threshold defined by the user return to step ii, else stop the ICM algorithm.

*Algorithm 1: ICM (Iterated Conditional Modes) algorithm*

Although the rapidity of this algorithm, its performance depends heavily on the initialization, also the course order of these superpixels has an influence on the convergence. In our case we use a simple thresholding algorithm to initialize the {Y$_i$} as the {X$_i$} and a path according to the numbers of superpixels in the image.

b. *Algorithm of model decomposition (C.M. Bishop 2006)*

In general, this algorithm seeks to find the optimal configuration, it recursively applies for each variable $Y_i$ unfixed to determine its value corresponds to the maximum likelihood of {P (T|$Y_i$=$C_j$),∀ j where T represents all the network variables}. We use the sum-product algorithm to calculate these probabilities which is based on a message passing architecture. This algorithm is presented formally as follows:

> **Input :**
>   - Image to segment, $[C_1... C_k]$: list of class centers.
>
> **Output :**
>   - Segmented image.
>
> **Algorithm :**
>
> While (exists $\{Y_i\}$ not yet fixed ) do
>
> 1. Select on variable $Y_i$ not yet fixed
> 2. For each class $C_j$, calculate P ($T|Y_i=C_j$) with the sum-product algorithm where T represents all fixed network variables.
> 3. Research the class with the maximum probability.
> 4. We retain the class $C_{max}$, find in step 3) for the superpixel $Y_i$.
>
> End while

*Algorithm 2: Model decomposition algorithm*

In conclusion, in this section we present two approximate inference methods, the first is based on a deterministic method, have the ability to converge quickly. But, its results depend heavily of the initial configuration. Therefore, a good initialization algorithm must be used in this case we can apply the model decomposition algorithm that is a very powerful algorithm, but it expresses also a major gap in the assignment of $Y_i$ class although there are other unfixed superpixels it can be improved by the ICM algorithm, in conclusion the combination decomposition algorithm and ICM algorithm can be very powerful at least in our case, it can be modeled as the following (fig.10):

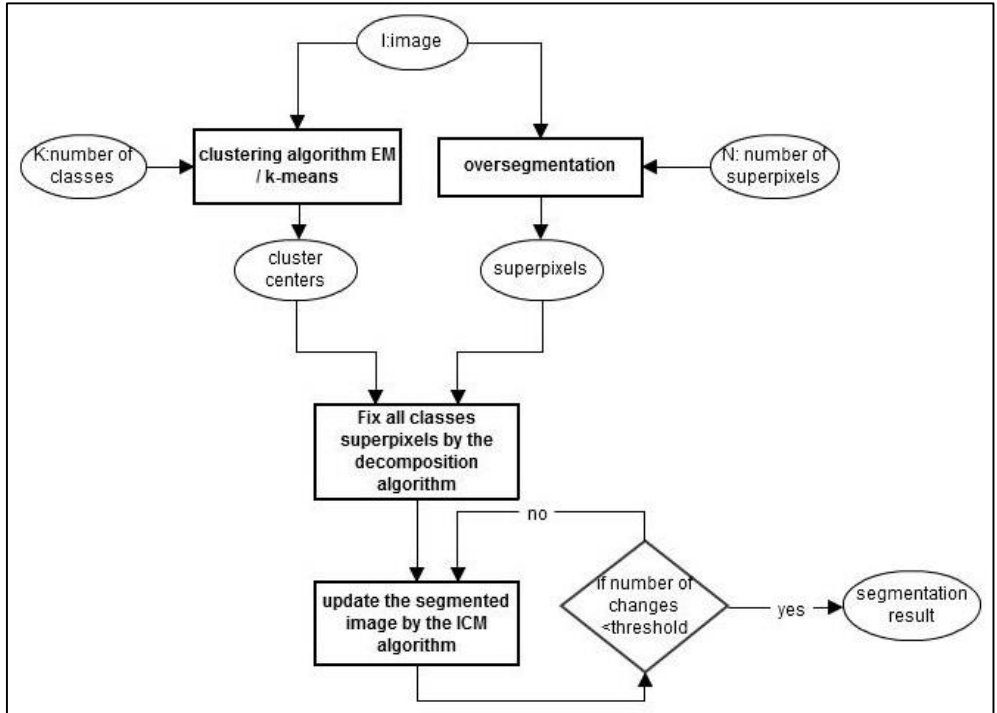
Fig.10 : General conduct inference

## 5. EXPERIMENTS AND RESULTS

The better of our approach is the ability to use many predicates at the same time in a simple way, using superpixel as a base element is also an advantage it reduces the processing time and exceeds the noise contained in images. In this section we present the advantages and the justification of over-segmentation algorithm used, and then we define the used predicates, and explaining the idea behind each it.

a.  Over-segmentation

In the latest research on image segmentation research, using superpixels instead of pixels is very frequent. Using superpixel can greatly reduce the search space images results and keeping the same performance as using pixels, for example, in a problem of labeling (L labels) the number of possible solutions for image of n pixels is $L^n$, in contrast if the image is represented by m superpixels where m << n, the number of solutions is $L^m$ where $L^m \ll L^n$. But the using of superpixels must check some properties:

- ❖ Each superpixel must overlap with a single object.
- ❖ The boundaries of a given object are a set of superpixels boundaries.
- ❖ The transition from pixels to superpixels should not reduce the performance of the application.
- ❖ The above properties should be obtained with little more superpixels possible.

In the selection of the over-segmentation algorithm, is sought an algorithm which checks the maximum of these properties. In this context we use The Entropy Rate Superpixel Segmentation (M. Liu 2011). In this algorithm, the over-segmentation is done according to an objective function composed of two terms, a term favors the melting of homogeneous clusters and another term slow and leads this fusion process to have a k superpixels where k is a parameter given by the user. This problem is considered as a graph partitioning problem, where the over-segmented image is represented by a weighted graph in each node represents a pixel of the image, and each arc represents the neighborhood relationship it expresses the degree of similarity. In conclusion, the problem is to partition the graph into k subgraphs maximizing the objective function using Greedy Heuristic algorithm (G. L. Nemhauser 1978)

*b.  Used predicates*

A predicate is a set of tests that can be applied to region $R_i$ composed by the superpixel $Y_i$ and its neighboring superpixels $R_i$ =SR1 ∪ $Y_i$ ∪ SR2. In our case we chose to use only two predicates {P1, P2} using the intensity information. In general, a predicate p seeks to express the homogeneity of $Y_i$ with SR1 and the contrast $Y_i$ over SR2.

- *P1 predicate*

   Whether:
   S: $Y_i$ superpixel intensity
   V:     the     standard     deviation     of     SR1     compared     to     S.
 L: the standard deviation of SR2 compared to S.
   P1 = True iff  (V< threshold_1) and (L> threshold_2).
threshold_1 and threshold_2 are two initialized parameters.

- *P2 predicate*

        Whether:
   S: $Y_i$ superpixel intensity.

SRV1= {V1, V2,…Vn}:  intensities set of SR1 superpixels.

SRV2= {L1, L2,…Lm}: intensities set of SR2 superpixels.

P2 = True **iff (**max (SRV1-S) <threshold_1) and (min (SRV2-S)>threshold_2)

threshold_1 and threshold_2 are two initialized parameters.

*c. Results*

In our experiments, we use our multi-layer model which is our main contribution. At First we evaluate all inferences algorithms studied to choose the best, then we will check the quality of this approach, as well as its shortcomings and advantages according to the results statistics, we finally propose possible improvements. Fig. 13 shows the segmentation results for data bases (http1 , http 2)

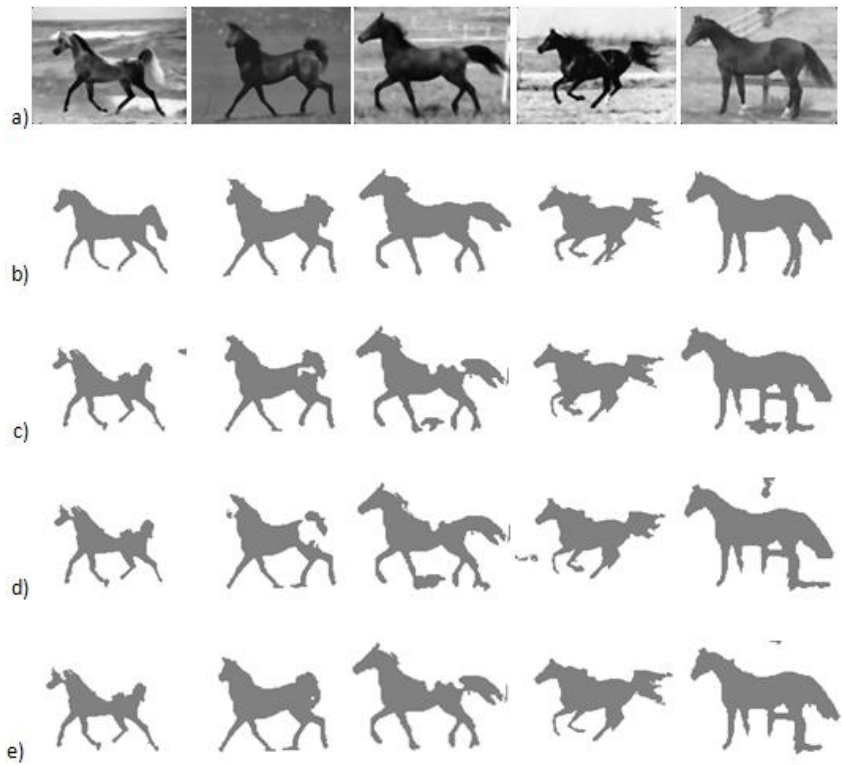

Fig. 13 a) Images b) ground truth, Segmentation by c) ICM d) Algorithm of model decomposition e) Algorithm of model decomposition + ICM

*Table 4 : Segmentation consistency according to inference algorithm used.*

| Inference algorithm | Number of processed images | Average processing time (sec) | Segmentation efficiency |
|---|---|---|---|
| ICM | 30 | 1.5 | 92.7% |
| Algorithm of model decomposition | 30 | 1.5 | 90.8% |
| Algorithm of model decomposition + ICM | 30 | 2 | **93.9%** |

We conclude that the combination algorithm of model decomposition and ICM algorithm express good performance so we will use in the sequel.

*Table 5: The Comparison of Our Approach with Several Related Works for Segmenting (Weizmann Data Set)*

| Method | Segmentation efficiency |
|---|---|
| Borenstein et al. (E. Borenstein 2006) (grey scale) | 93% |
| Winn et al. (] J. Winn 2005) (grey scale) | 93% |
| Lei Zhang, et Qiang Ji (Lei Zhang 2010) (color) | 93.7% |
| Our approach (grey scale) | **93.9 %** |

In the next section we will present the results of our approach in the case of multi-class segmentation on a few images BSDB.

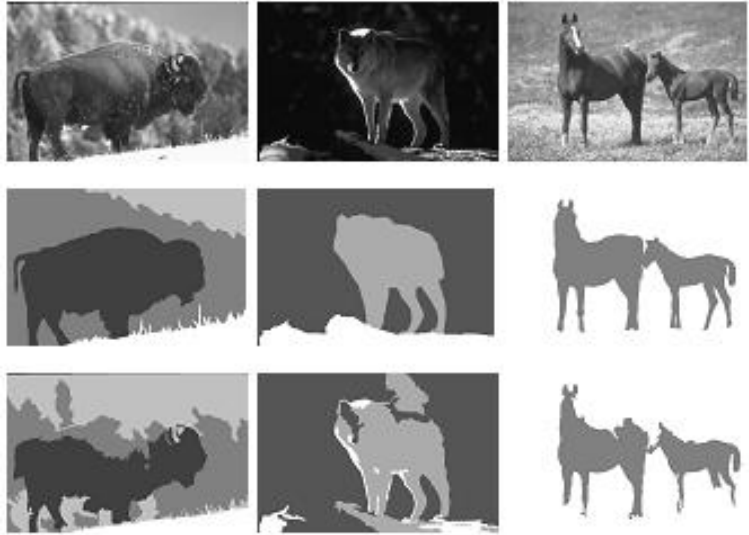

Figure 14 : Multi-classes Segmentation

our approach express also a strong performance in the multi-classes.

*Table 6 : multi-classes Segmentation consistency.*

| Images | Image size | Average processing time (sec) | Segmentation efficiency |
|---|---|---|---|
| Image1 | 481*321 | 7.23 | 70.7% |
| Image2 | 481*321 | 6.7 | 75.5% |
| Image3 | 481*321 | 5.28 | 85.9% |

Figures 15 and 16 show the intermediate results of segmentation in the case of 2 and 3 classes.

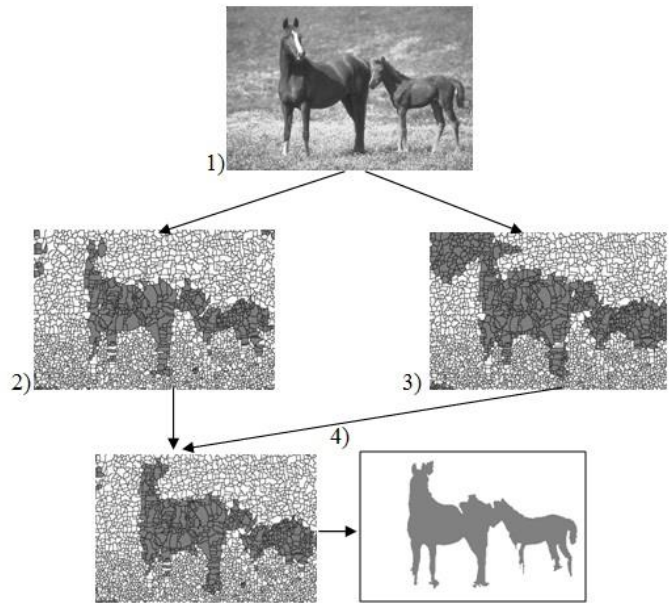

Figure 15 : 1-image gray level ; 2-result of first part of network ; 3-result of second part of network 4- result of unified network

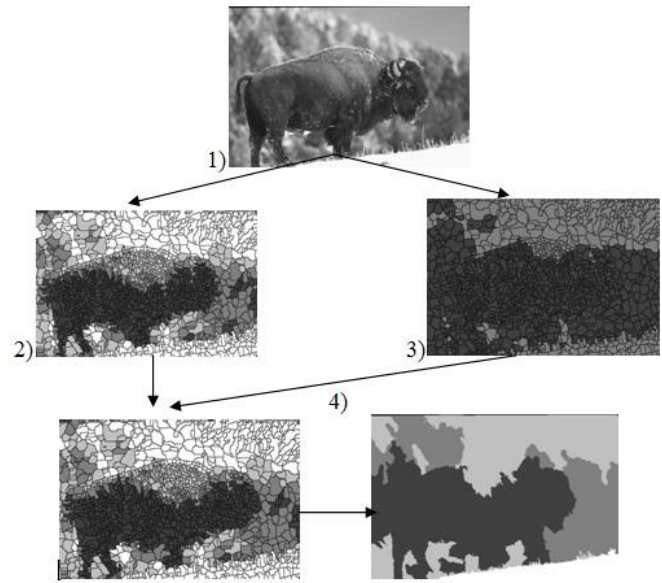

Figure 16 : 1-image gray level ; 2-result of first part of network ; 3-result of second part of network 4- result of unified network

*d. Segmentation of old documents*

For a better validation of our approach, we test it with ancient documents from the National Library of Tunisia (Fig.17) shows the results of binarization of the image.

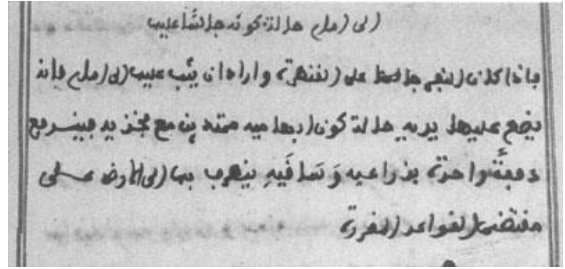

Fig. 17: Image of ancient documents from the National Library of Tunisia

To showcase our approach, we will compare our model with those known and recognized in the field of document binarization. In looking at the literature, we note that recent work and confirmed in the field are those of Otsu (Otsu N 1979) of Niblack (W. Niblack 1986) and Sauvola (J. Sauvola 1997). Figure 18 and 19 showsthe results of binarization of the image in Figure 17 by the 3 three models.

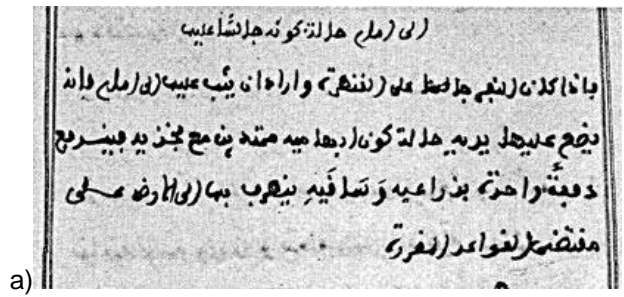

a)

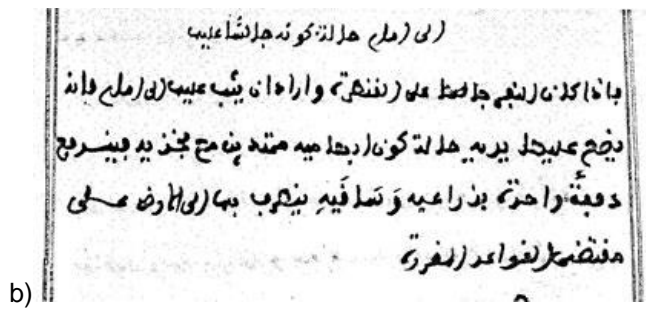

b)

Fig. 18 a) Image Segmentation by a) Niblack[28] b) Sauvola [29]

For this type of image, the used predicates are not specified to judge the segmentation quality, but the results of our approach show good performances compared with other three models. In conclusion, our approach expresses good results, also it discover a very competitive advantage which is the processing time. In this context we study in the next section the behavior of our approach in terms of processing time and efficiency of segmentation compared to the number of superpixels.

a)

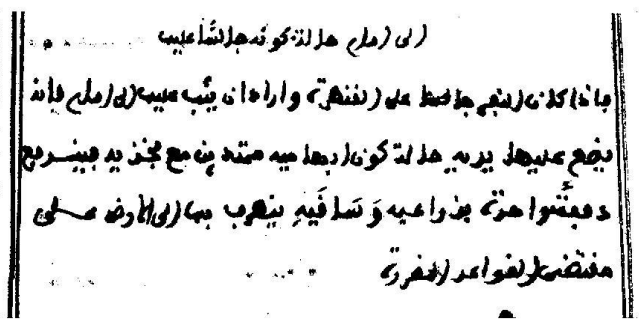

b)

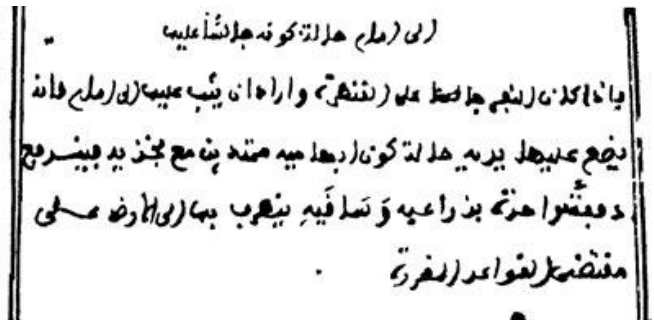

Fig. 19 a) Image Segmentation by a) Otsu b) Proposed approach multi layers

Although the predicates used are not specified for this field segmentation of documents to judge the quality of the segmented image, results of our approach shows good performance compared with other three models.

In conclusion, our approach expresses strong performance point of view but also results she discovers a very competitive advantage which is the processing time, this time mainly depends number of superpixels. In this context we will study in the next section the behavior of our approach in terms of time and processing efficiency of segmentation compared to the number of superpixels.

e. *Error analyzis*

In conclusion, our approach expresses strong performance point of view but also results she discovers a very competitive advantage which is the processing time, this time mainly depends number of superpixels. In this context we will study in the next section the behavior of our approach in terms of time and processing efficiency of segmentation compared to the number of superpixels.

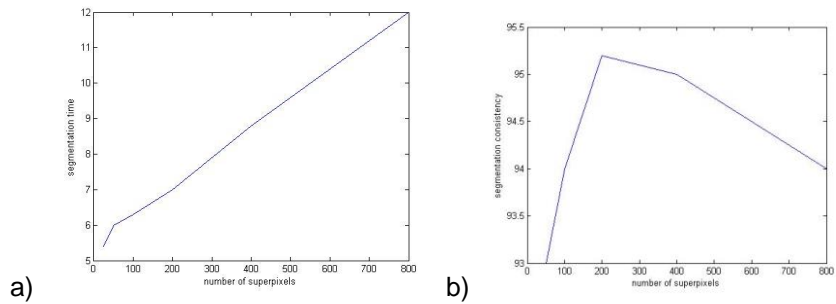

Fig.20: Behavior of a) Processing time b) Segmentation consistency relative to the number of superpixels.

From these curves (fig. 20) we see that the complexity of our approach is O (n) where n is the number of superpixels, we also note that the increase in the number of superpixels does not increase the effectiveness. So for each image there are an ideal number of superpixels, this is a defect of our approach. In our model, to calculate the probabilities $P(Y_i | X_i)$ and $P(R_i | pa(R_i))$, we use the normal distribution (µ, σ) where µ represents the class center and σ the standard deviation of this class. In the next section we will therefore study the effectiveness of segmentation with respect to this parameter µ.

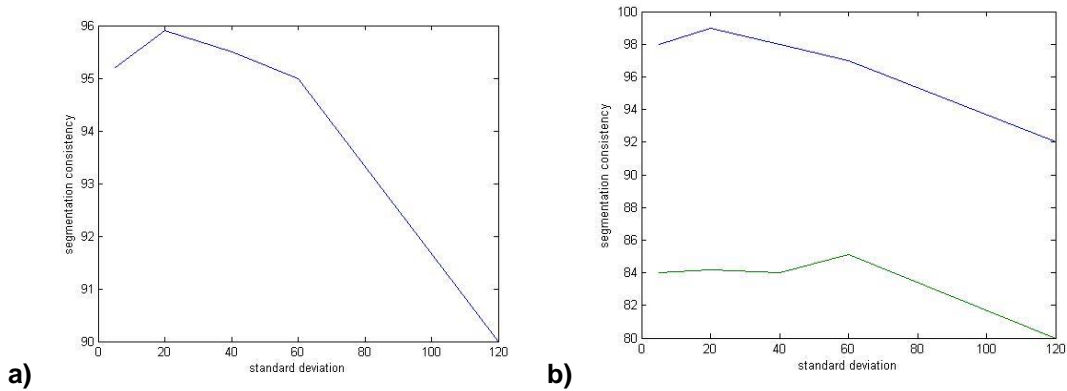

Fig.21: segmentation consistency a) of all segmented image b) of each class relative to the standard deviation.

From these curves (fig. 21) we see that the efficiency of segmentation depends of initialized standard deviation, you can even notice that for each class there is a perfect deviation different, the random initialization of this parameter for all classes in our case represents a defect in our approach.

## 6. CONCLUSION

Bayesian networks offer great potential for modeling data from images in a causal way. At first, the using of these models was restricted to decision support systems, with the success, they are now used in image processing. In this context we propose in this work two segmentation approaches using these oriented graphical models.

The second model is a multi-layer, which represents our main contribution. It is based on two phases: over-segmentation phase and inference phase based on two approximate inference methods, the first use ICM algorithm (Iterated Conditional Mode) which is widely used in the Markov model and the second is a recursive algorithm based on max-prodcut algorithm, very popular in the recent work of Bayesian networks segmentation called algorithm of model decomposition. From the results statistics, we conclude that the combination algorithm of model decomposition and ICM algorithm are complementary at least in our approach, the comparison with other models already studied in the literature demonstrates the robustness of our model. The better of our approach is the ability to use many predicates at the same time in a simple way, these predicates can express area or contour approach, they can also use the intensity or texture information. Using superpixel as a base element is also an advantage it reduces the processing time and exceeds the noise contained in images. In conclusion, our approach expresses good results, also it discover a very competitive advantage which is the processing time. This time depends mainly on the number of superpixels. In addition, our model can be greatly improved, especially by adding other predicates evaluation based on texture and shape.

# References

Brunel, Nicolas and Pieczynski, Wojciech, "Unsupervised signal restoration using hidden Markov chains with copulas" Signal Processing Vol. 85 pp. 2304—2315

K.E. Avrachenkov, E. Sanchez. Fuzzy Markov chains. *Fuzzy Optimization and Decision Making*, 1(2) : 143-159, 2002.

Lei Zhang and Qiang Ji, Image Segmentation with a Unified Graphical Model, IEEE Transactions on Pattern Analysis and Machine Intelligence, Issue 8, Vol.32, pages 1406-1425, 2010.

E.N. Mortensen and J. Jia, "Real-Time Semi-Automatic Segmentation Using a Bayesian Network," Proc. IEEE CS Conf. Computer Vision and Pattern Recognition, pp. 1007-1014, 2006.

Kittipat, Kampa, Duangmanee Putthividhya and Jose C. Principe, "Irregular Tree-Structured Bayesian Network for Image Segmentation.", IEEE International Workshop on Machine Learning for Signal Processing September 18-21, 2011.

Costas Panagiotakis, Ilias Grinias, and Georgeios Tziritas "Natural Image Segmentaion Based on Tree Equipartition, Bayesian Flooding and Region Merging", *IEEE Transactions on Image Processing*,Vol. 20, No. 8, August 2011.

F. Liu, D. Xu, C. Yuan, and W. Kerwin, "Image segmentation based on Bayesian network-Markov random field model and its application on in vivo plaque composition," in *Int. Symp. Biomed. Imag.*,2006, pp. 141 –144.

X. Feng, C. Williams, and S. Felderhof, "Combining Belief Networks and Neural Networks for Scene Segmentation," IEEE Trans. Pattern Analysis and Machine Intelligence, vol. 24, no. 4, pp. 467-483, Apr. 2002.

S. Zheng, Z. Tu, and A. Yuille, "Detecting Object Boundaries Using Low-, Mid-, and High-Level Information," Proc. IEEE CS Conf. Computer Vision and Pattern Recognition, pp. 1-8, 2007.

S. Khan and M. Shah, "Object Based Segmentation of Video Using Color, Motion and Spatial Information," Proc. IEEE CS Conf. Computer Vision and Pattern Recognition, vol. 2, pp. 746-751, 2001.

Z. Tu, X. Chen, A.L. Yuille, and S.-C. Zhu, "Image Parsing: Unifying Segmentaion, Detection, and Recognition," Proc. IEEE Int'l Conf. Computer Vision, pp. 18-25, 2003.

Z. Tu, C. Narr, P. Dollar, I. Dinov, P. Thompson, and A. Toga, "Brain Anatomical Structure Segmentation by Hybrid Discriminative/ Generative Models," IEEE Trans. Medical Imaging, vol. 27, no. 4, pp. 495-508, Apr. 2008.

S. Horowitz et T. Pavlidis: Picture segmentation by a directed split-and-merge procedure. Rapport technique, Departement of Electrical Engineering, Princeton University, 1975.

S. Horowitz et T. Pavlidis : Picture segmentation by a tree traversal algorithm. Journal of The Association for Computing Machinery, 23(3):368–388, avril 1976.

F.V Jensen, "Bayesian network and decision graph", SpringerVerlag, USA, 2001.

P.Naim, P.H.Wuillemin, P.Leray, O.Pourret, A.Becker. Réseaux bayésiens, Eyrolles, Paris, 2005.

O.Francois. De l'identification de structure de réseaux bayésiens à la reconnaissance de formes à partir d'informations complètes ou incomplètes. Thèse de doctorat, Institut National des Sciences Appliquées de Rouen, 2006.

N.B.Amor, S.Benferhat, Z.Elouedi. Réseaux bayésiens naïfs et arbres de décision dans les systèmes de détection d'intrusions. RSTI-TSI. Volume 25 - n°2/2006, pages 167 à 196.

Fabien SAIZENSTEIN et Wojciech PIECZYNSKI "On the Choice of Statistical Image Segmentation Method" Traitement du Signal 1998 — Volume 15 - n°2

M. Liu, O. Tuzel, S. Ramalingam, and R. Chellappa, "Entropy rate superpixel segmentation," in *CVPR*, 2011, pp. 2097–2104.

Besag, J. (1986). ``On the statistical analysis of dirty pictures'' (with discussions). *Journal of the Royal Statistical Society, Series B*, 48:259--302.

C.M. Bishop, Pattern Recognition and Machine Learning. Springer, 2006.

G. L. Nemhauser, L. A. Wolsey, and M. L. Fisher. An analysis of the approximations for maximizing submodular set functions. *Mathematical Programming*, pages 265–294, 1978.

E. Borenstein and J. Malik, "Shape Guided Object Segmentation," Proc. IEEE CS Conf. Computer Vision and Pattern Recognition, pp. 969-976, 2006.

J. Winn and N. Jojic, "Locus: Learning Object Classes with Unsupervised Segmentation," Proc. IEEE Int'l Conf. Computer Vision, pp. 756-763, 2005.

W. Niblack, An Introduction to Image Processing, Prentice-Hall, Englewood Cliffs, NJ, 1986.

J. Sauvola, T. Sepp¨anen, S. Haapakoski, and M. Pietik¨ainen, "Adaptive document binarization," in ICDAR '97: Proceedings of the 4th International Conference on Document Analysis and Recognition. 1997, pp. 147–152, IEEE Computer Society.

Otsu N. "A Threshold Selection Method from Gray-level Histograms," IEEE Trans. Syst. Man Cybern, 1979, 9: 62-66.

[http 1] : http://www.eecs.berkeley.edu/Research/Projects/CS/vision/bsds/.

[http 2] http://jamie.shotton.org/work/data.html